\def\year{2018}\relax
\documentclass[letterpaper]{article} 
\usepackage{graphicx}
\usepackage{aaai18}  
\usepackage{times}  
\usepackage{helvet}  
\usepackage{courier}  
\usepackage{url}  
\usepackage{graphicx}  
\usepackage{multirow}
\usepackage{amsmath}
\newcommand{\cev}[1]{\reflectbox{\ensuremath{\vec{\reflectbox{\ensuremath{#1}}}}}}
\begin{document}
%
\title{Improved English to Russian Translation by Neural Suffix Prediction}
\author{Kai Song\textsuperscript{1,2},
Yue Zhang\textsuperscript{3},
Min Zhang\textsuperscript{1},
Weihua Luo\textsuperscript{2}\\
\textsuperscript{1} Soochow University, Suzhou, China\\
\textsuperscript{2} Alibaba Group, Hangzhou, China\\
\textsuperscript{3} Singapore University of Technology and Design, Singapore\\
\{songkai.sk, weihua.luowh\}@alibaba-inc.com\\
yue\_zhang@sutd.edu.sg, minzhang@suda.edu.cn\\
}
\maketitle
\begin{abstract}
Neural machine translation (NMT) suffers a performance deficiency when a limited vocabulary fails to cover the source or target side adequately, which happens frequently when dealing with morphologically rich languages. To address this problem, previous work focused on adjusting translation granularity or expanding the vocabulary size. However, morphological information is relatively under-considered in NMT architectures, which may further improve translation quality. We propose a novel method, which can not only reduce data sparsity but also model morphology through a simple but effective mechanism. By predicting the stem and suffix separately during decoding, our system achieves an improvement of up to 1.98 BLEU compared with previous work on English to Russian translation. Our method is orthogonal to different NMT architectures and stably gains improvements on various domains.
\end{abstract}

\section{Introduction}
Neural machine translation (NMT) \cite{bahdanau2014neural} has shown better performance compared with statistic machine translation \cite{Zens2002Phrase}. Such methods encode a source sentence into hidden states and generate target words sequentially by calculating a probability distribution on the target-side vocabulary. Most NMT systems limit target side vocabulary to a fixed size, considering the limit of graphics memory size and high computing complexity when predicting a word over the whole target side vocabulary (e.g., 30K or 50K). In addition, a larger target-side vocabulary can also make the prediction task more difficult. Word-level NMT systems suffer the problem of out of vocabulary (OOV) words, particularly for morphologically rich languages. For example, English to Russian machine translation faces a big challenge due to rich morphology of Russian words, which leads to much more OOV words than some other languages. Typically a specific tag is used to represent all OOV words, which is then translated during a post process \cite{Luong2014Addressing}. This can be harmful to the translation quality.

There has been several methods to address this problem. Some focused on translation granularity (\citeauthor{sennrich2015neural}, \citeyear{sennrich2015neural}; \citeauthor{lee2016fully}, \citeyear{lee2016fully}; \citeauthor{luong2016achieving}, \citeyear{luong2016achieving}), while others (\citeauthor{jean2014using}, \citeyear{jean2014using}; \citeauthor{mi2016vocabulary}, \citeyear{mi2016vocabulary}) effectively expand target side vocabulary. However, though those methods can avoid OOV, none of them has explicitly modeled the target side morphology. When dealing with language pairs such as English-Russian, the number of different target side words is large due to the rich suffixes in Russian. The above methods are limited in distinguishing one suffix from another.

\begin{figure*}[t]
\center
\includegraphics[width=0.9\textwidth]{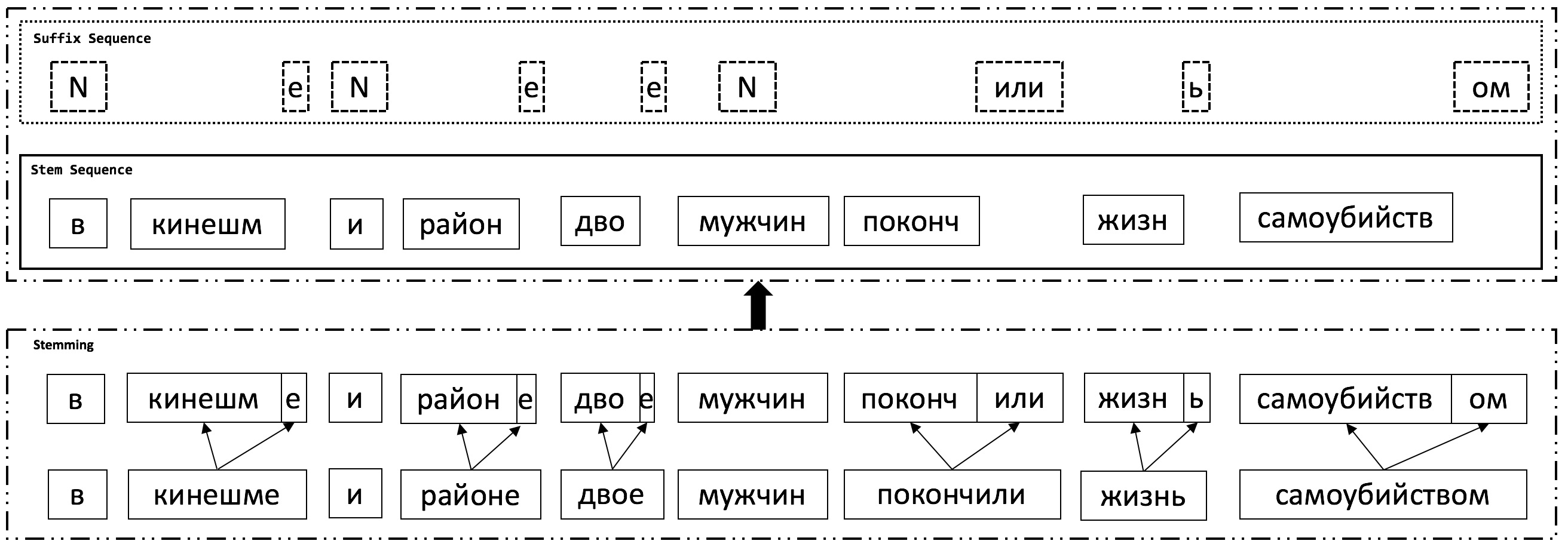}
\caption{Russian word sequence to stem sequence and suffix sequence, ``N'' is a special tag used in suffix sequence, which means ``no suffix'' for corresponding stem.}
\label{fig:stemsequence}
\end{figure*}

Since the total number of different stems in a morphologically rich language is much less than the number of words, a natural perspective to make a better translation on a morphologically-rich target-side language is to model stems and suffixes separately. We design a simple method, which takes a two-step approach for the decoder. In particular, stem is first generated at each decoding step, before suffix is predicted. Two types of target side sequences are used during training, namely stem sequence and suffix sequence, which are extracted from the original target side word sequence, as shown in Figure \ref{fig:stemsequence}. Sparsity is relieved since the number of stem types is much smaller than word types, and suffix types can be as small as several hundreds. Another advantage of this structure is that during the prediction of suffix, the previously generated stem sequence can be considered, which can further improve the accuracy of suffix prediction.

We empirically study this method and compare it with previous work on reducing OOV rates (\citeauthor{sennrich2015neural}, \citeyear{sennrich2015neural}; \citeauthor{lee2016fully}, \citeyear{lee2016fully}). Results show that our method gives significant improvement on the English to Russian translation task on two different domains and two popular NMT architectures. We also verify our method on training data consisting of 50M bilingual sentences, which proves that this method works effectively on large-scale corpora.

\begin{figure}[t]
\center
\includegraphics[height=0.18\textwidth]{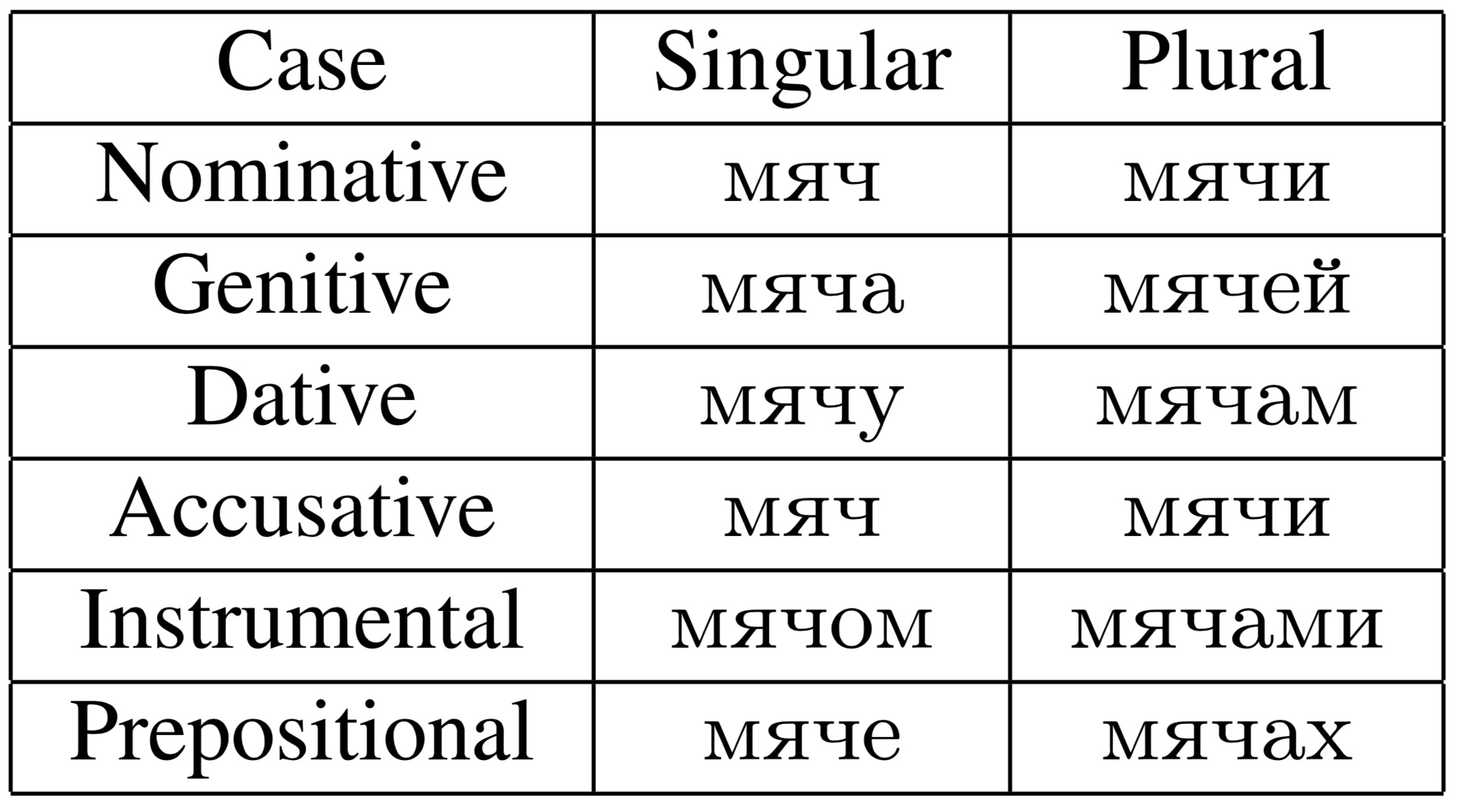}
\caption{Different forms of the word ``ball''.}
\label{fig:russian_words}
\end{figure}

\section{Related Work}
\subsection{Translation Granularity}
Subword based \cite{sennrich2015neural} and character-based (\citeauthor{lee2016fully}, \citeyear{lee2016fully}; \citeauthor{luong2016achieving}, \citeyear{luong2016achieving}) NMT are the two directions of adjusting translation granularity, which can be helpful to our problem.

In \citeauthor{sennrich2015neural} (\citeyear{sennrich2015neural})'s work, commonly appearing words remain unchanged, while others are segmented into several subword units, which are from a fixed set. Both source and target side sentences can be changed into subword sequences. More specifically, some rare words are split into and represent as some more frequent units, base on a data compression technique, namely Byte Pair Encoding (BPE). The vocabulary built on common words and these frequent subword units can successfully improve the coverage of training data. In fact, a fixed size vocabulary can cover all the training data as long as the granularity of subword units is small enough. The main limitation of this method is the absence of morphology boundary. Some subword units may not be a word suffix which can represent a morphological meaning, and the subword units are treated in the same way as complete words. Subword units and complete words are predicted during a same sequence generation procedure. This may lead to two problems:
\begin{itemize}
\item The sequence length can increase, especially on a morphologically rich language, which can lead to low NMT performance.
\item A subword unit cannot represent a linguistic unit, and suffix is not modeled explicitly.
\end{itemize}

\citeauthor{luong2016achieving} (\citeyear{luong2016achieving}) proposed a hybrid architecture to deal with the OOV words in source side and any generated unknown tag in the target side. In their system, any OOV words on the source side are encoded at the character level, and if an unknown tag is predicted during decoding, another LSTM will be used to generate a sequence of target-side characters, which will be used as the replacement of the target side unknown word for the translation of a source OOV. However, their model may not work well when the target side is morphologically rich and the source side is not, because their hybrid network on the target side will only be used when an unknown tag is generated, which is always corresponding to a source unknown word. If most of the source side tokens are covered by the source vocabulary, the hybrid network may not have advantage on a morphologically rich target side language.

In \citeauthor{lee2016fully} (\citeyear{lee2016fully})'s work, source side and target side sequence are all character-based, which eliminates OOV on the source side, and can generate any target side word theoretically. Character-based NMT may potentially improve the translation accuracy of morphologically rich language on the source side, but the training and decoding latency increase linearly with the sequence length, which is several times to the original word based NMT. Another disadvantage of character-based NMT is that character embedding lost the ability to represent a linguistic unit. Long-distance dependences are more difficult to be modeled in a character-based NMT. \citeauthor{lee2016fully} (\citeyear{lee2016fully}) use convolutional and pooling layers on the source side to make the source sequence shorter. However, the target side sequence remains much longer than the original word sequence, and suffix boundary of the target side is not specifically considered in their model. This work may more helpful if a morphologically rich language is on the source side, but it is not designed to overcome the problem brought by a morphologically rich target side language.

There is another way which can effectively reduce target-side OOV. Both \citeauthor{jean2014using} (\citeyear{jean2014using}) and \citeauthor{mi2016vocabulary} (\citeyear{mi2016vocabulary}) use a large target-side vocabulary. To overcome the problem of GPU memory limitation and increasing computational complexity, instead of the original vocabulary, a selected subset is actually used both during the training and decoding time. Their model can generate any of the words in the large vocabulary, but data sparsity still remains, the low frequent words in the training data is not fully trained.
\subsection{Morphology and MT}
Previous work considered morphological information for both SMT and NMT. \citeauthor{koehn2007factored} (\citeyear{koehn2007factored}) proposed an effective way to integrate word-level annotation in SMT, which can be morphological, syntactic, or semantic. Morphological information can be utilized not only on source side, but also the target side. Although these annotation can help to improve the translation procedure, data sparsity still exists. \citeauthor{chahuneau2013translating} (\citeyear{chahuneau2013translating}) decompose the process of translating a word into two steps. Firstly a stem is produced, then a feature-rich discriminative model selects an appropriate inflection for the stem. Target-side morphological features and source-side context features are utilized in their inflection prediction model.

\citeauthor{tran2015distributed} (\citeyear{tran2015distributed}) use distributed representations for words and soft morphological tags in their neural inflection model, which can effectively reduce lexical sparsity, leading to less morphological ambiguity. This is the first try of modeling inflection through a neural method, integrated in a SMT architecture.

For NMT, \citeauthor{sennrich2016linguistic} (\citeyear{sennrich2016linguistic}) make use of various source side features (such as morphological features, part-of-speech tags, and syntactic dependency labels) to enhance encoding in NMT. This is the first time morphological information is leveraged in NMT architecture. Target-side morphology is not considered in their work. \citeauthor{Ale2017Modeling} (\citeyear{Ale2017Modeling}) predict a sequence of interleaving morphological tags and lemmas, followed by a morphological generator. They used a external model to synthesize words given tags and lemmas. Our method is the first to explicitly consider the generation of morphological suffixes within a neural translation model. Our work is motivated by a line of work that generates morphology during text generation (\citeauthor{Toutanova2010Applying}, \citeyear{Toutanova2010Applying}; \citeauthor{song2014joint}, \citeyear{song2014joint}; \citeauthor{tran2015distributed}, \citeyear{tran2015distributed}).
\section{Background}
\subsection{Russian Morphology and Stemming}
\textbf{Morphology} Russian has rich morphology, which includes number (singular or plural), case (nominative, accusative etc.), gender (feminine, masculine or neuter) and tense mood. Figure \ref{fig:russian_words} shows one example for Russian. A noun word ``ball'' is always masculine, but the suffix differs when the case and number changes, resulting in 10 different forms. Some other nouns can be feminine or neuter, and their adjectives will agree with them. Both adjectives and verbs have different forms according to their case, tense mood and the form of words they modify. Such morphological changes bring a challenge to machine translation task. 

\textbf{Stemming} A Russian word can be split into two parts, namely the stem and the suffix. Suffix contains morphological information of a Russian word, including gender, number and case etc. In this paper, we use a deterministic rule-based stemmer to obtain stem and suffix for a Russian word. The process of stemming is shown in Figure \ref{fig:stemsequence}.

\subsection{Neural Machine Translation Baselines}
We experiment with two different types of Neural Machine Translation (NMT) systems, one using a recurrent encoder-decoder structure \cite{bahdanau2014neural}, the other leveraging the attention mechanism on the encoder \cite{DBLP:journals/corr/VaswaniSPUJGKP17}.

\textbf{Recurrent Neural Network Based NMT} We use an encoder-decoder network proposed by \citeauthor{cho2014properties} (\citeyear{cho2014properties}). The encoder uses a bi-directional recurrent neural network (RNN) to encode the source sentence, the decoder uses a uni-directional RNN to predict the target translation. Formally, the source sentence can be expressed as $\textbf{x} = (x_1, ..., x_m)$, where $m$ is the length of the sentence. It is encoded into a sequence of hidden states $ \textbf{h} = (h_1, ..., h_m)$, each $h_i$ is the result of a concat operation on a forward (left-to-right) hidden state $\vec h_i$ and a backword (right-to-left) hidden state $\cev h_i$:
\begin{equation}
\label{source-hidden}	
h_i = [\vec h_i;\cev h_i],
\end{equation}
\begin{equation}
\label{forward-src-hidden}
\vec h_i = f(\vec h_{i-1}, x_i),
\end{equation}
\begin{equation}
\label{backward-src-hidden}
\cev h_i = f(\cev h_{i+1}, x_i)
\end{equation}

$f$ is a variation of LSTM \cite{hochreiter1997long}, namely Gated Recurrent Unit (GRU) \cite{cho2014learning}:
\begin{equation}
\label{z-gate}
z_i = \sigma (\textbf{W}_z * [h_{i-1}, x_i]),
\end{equation}
\begin{equation}
\label{r-gate}
r_i = \sigma (\textbf{W}_r * [h_{i-1}, x_i]),
\end{equation}
\begin{equation}
\label{h-gate}
\tilde{h_i} = {\it tanh} (\textbf{W} * [r_i * h_{i-1}, x_i]),
\end{equation}
\begin{equation}
\label{gru}
h_i = (1 - z_i) * h_{i-1} + z_i * \tilde{h_i}
\end{equation}
where $\textbf{W}_z$, $\textbf{W}_r$, $\textbf{W}$ are weight matrices which are learned.

During decoding, at each time step $t$, an attention probability $\alpha_{tj}$ to the source word $x_j$ is first calculated by:
\begin{equation}
\label{attention}
\alpha_{tj}=\frac{\exp(e_{tj})}{\sum_{k=1}^{m}\exp(e_{tk})}
\end{equation}
and
\begin{equation}
\label{energy}
e_{tj}=a(S_{t-1}, h_j)
\end{equation}
is an attention model that gives a probability distribution on source words $(x_1, ... , x_m)$, which indicates how much the source word $x_j$ is considered during the decoding step $t$ to generate target side word $y_t$. The attention layer $a$ can be as simple as a feed-forward network. $C_t$ is a weighted sum of the encoding hidden state at each position of input sentence:
\begin{equation}
\label{src-context}
C_t=\sum_{j=1}^{m}(\alpha_{tj}*h_j),
\end{equation}
$C_t$ is then fed into a feed-forward network together with previous target word embedding $y_{t-1}$ and the current decoding hidden state ${S_t}$ to generate the output intermediate state $O_t$:
\begin{equation}
\label{output-layer}
O_t=g(y_{t-1}, S_t, C_t),
\end{equation}
and
\begin{equation}
\label{decoding-hidden}
S_t=f(y_{t-1}, S_{t-1}, C_t),
\end{equation}
where $f$ is GRU, which is mentioned before. The output intermediate state $O_t$ is then used to predict the current target word by generating a probability distribution on target side vocabulary. In our implementation, maxout \cite{goodfellow2013maxout} mechanism is used in both training and decoding. Dropout \cite{srivastava2014dropout} is used in training time.

\textbf{Transformer} \cite{DBLP:journals/corr/VaswaniSPUJGKP17} is a recently proposed model for sequence to sequence tasks. It discards the RNN structure for building the encoder and decoder blocks. Instead, only the attention mechanism is used to calculate the source and target hidden states.

The encoder is composed of stacked neural layers. In particularly, for the time step $i$ in layer $j$, the hidden state $h_{ij}$ is calculated as follows: First, a self-attention sub-layer is employed to encode the context. For this end, the hidden states in the previous layer are projected into a tuple of queries($Q$), keys($K$) and values($V$), where $g$ in the following function denotes a feed forward layer:
\begin{equation}
\label{QKV}
    Q_{(i(j-1)},K_{i(j-1)},V_{i(j-1)} = g(h_{i(j-1)})
\end{equation}

Then attention weights are computed as scaled dot product between current query and all keys, normalized with a softmax function. After that, the context vector is represented as weighted sum of the values projected from hidden states in the previous layer. The hidden state in the previous layer and the context vector are then connected by residual connection, followed by a layer normalization function \cite{ba2016layer}, to produce a candidate hidden state $h_{ij}^{'}$. Finally, another sub-layer including a feed forward layer, followed by another residual connection and layer normalization, are used to obtain the hidden state $h_{ij}$:
\begin{equation}
\label{hidden_state}
    h_{ij} = {\rm LayerNorm}(h_{ij}^{'} + g(h_{ij}^{'}))
\end{equation}

The decoder is also composed of stacked layers. The hidden states are calculated in a similar way, except for the following two differences: First, only those target positions before the current one are used to calculate the target side self-attention. Second, attention is applied in both target-to-target and target-to-source. The target-to-source attention sub-layer is inserted between the target self-attention sub-layer and the feed-forward sub-layer. Different from self-attention, the queries($Q$) are projected from target hidden states in the previous layer, and the keys($K$) and values($V$) are projected from the source hidden states in the last layer.

The rest of the calculation is exactly the same with self-attention. Compared to RNN based sequence to sequence models, transformer allows significantly more parallelization, since all the hidden states in the same layer can be calculated simultaneously, whereas the hidden states in RNN can only be calculated sequentially from left to right. In consideration of translation quality, \citeauthor{DBLP:journals/corr/VaswaniSPUJGKP17} (\citeyear{DBLP:journals/corr/VaswaniSPUJGKP17}) use multi-head attention instead of single-head attention as mentioned above, and positional encoding is also used to compensate the missing of position information in this model.

\begin{figure}[t]
\center
\includegraphics[height=0.4\textheight]{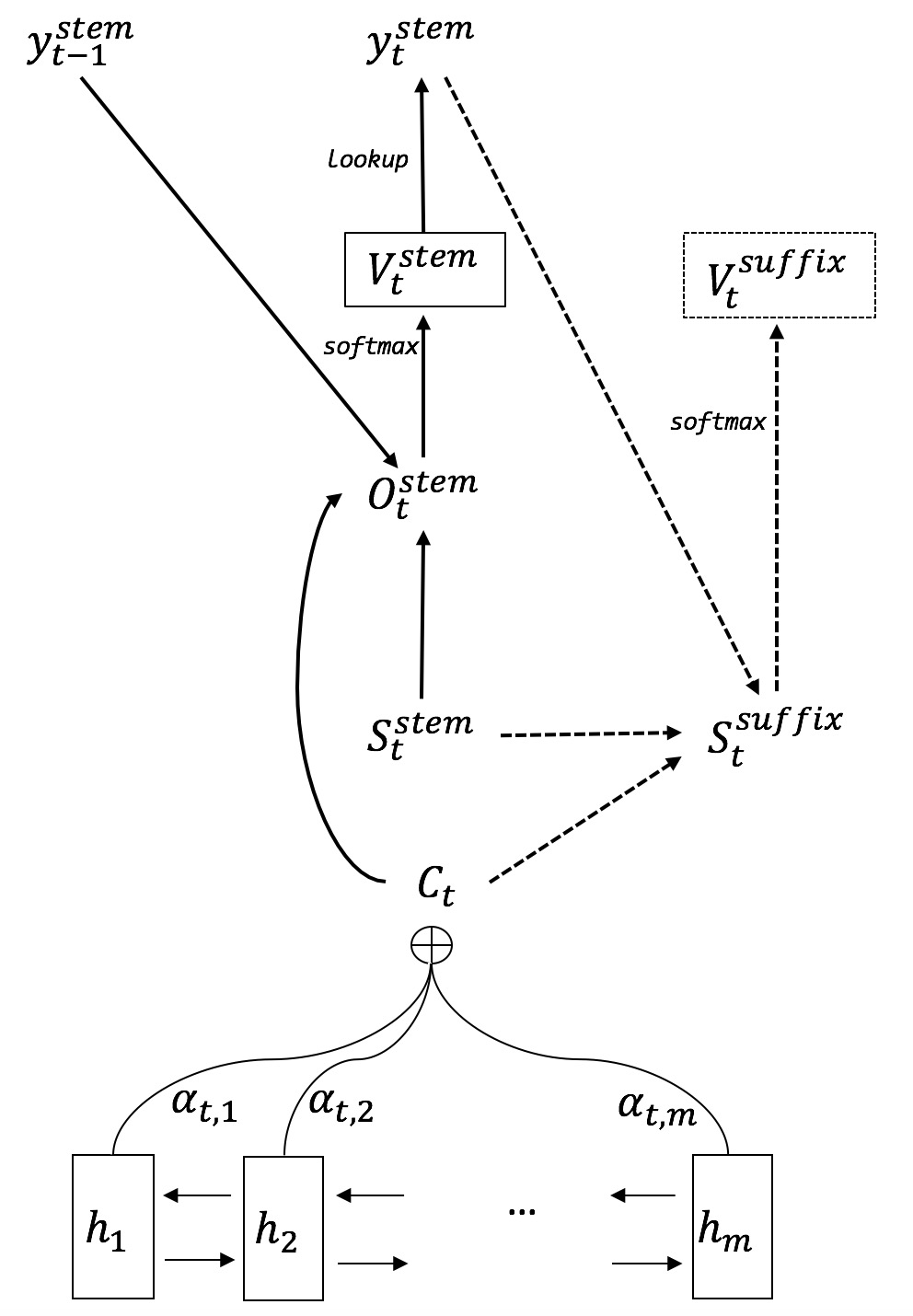}
\caption{Improved rnn-based NMT architecture.}
\label{fig:suffix_v1_rnn}
\end{figure}

\section{Target-Side Suffix Prediction}
We take a two-step approach for the decoder, yielding a stem at each time step before predicting the suffix of the stem. Since we only make use of source hidden states, target hidden states, target to source attention weights and target predicted tokens, these are universal in all sequence to sequence models, our method can be implemented into any of these models.

Figure \ref{fig:suffix_v1_rnn} shows a more detailed procedure. Decoding target stems is exactly the same as decoding target words in normal sequence to sequence model, which is predicted through a {\rm softmax} layer based on the target output layer. All we need is to replace target words with target stems:
\begin{equation}
\label{soft-stem}
 \begin{split}
 p(y_t^{\it stem}|&y_{1}^{stem},...,y_{t-1}^{stem},\textbf{x}) = \\
 &{\rm softmax}(O_t^{stem} * \textbf {W}^{stem}),
 \end{split}
\end{equation}
where $\textbf {W}^{stem}$ is a weight matrix to transfer the output layer $O_t^{stem}$ from a dimension of hidden size to target side vocabulary size. $S_t^{\it stem}$ is target side hidden state at time step $t$ when generating the stem. $O_t^{stem}$ is the output state:
\begin{equation}
\label{out-stem}
 \begin{split}
 O_t^{stem} = g(y_{t-1}^{stem}, S_t^{stem}, C_t),
 \end{split}
\end{equation}
$g$ is a single layer feed-forward neural network. 

After the prediction of $y_t^{\it stem}$, the target suffix $y_t^{\it suffix}$ on decoding step $t$ is immediately predicted from the target suffix hidden state $S_t^{\it suffix}$:
\begin{equation}
\label{soft-suffix}	
 \begin{split}
 p(y_t^{\it suffix}|&y_{1}^{stem},...,y_{t}^{stem},\textbf{x}) = \\
 &{\rm softmax}(S_t^{\it suffix} * \textbf {W}^{\it suffix})
\end{split}
\end{equation}

$S_t^{\it suffix}$ is generated from a single layer feed-forward neural network by using the stem embedding $y_t^{\it stem}$, stem hidden state $S_t^{\it stem}$, and source context vector $C_t$:
\begin{equation}
\label{hidden-suffix}	
S_t^{\it suffix}=g(S_t^{\it stem}, y_t^{\it stem}, C_t)
\end{equation}

Since we consider that the attention degree towards each word in the source sequence is useful to the generation of suffix, the aligned source context is also used during the prediction of suffix. Note that the source context vector $C_t$ is shared between the generation of stem hidden state $S_t^{\it stem}$ and suffix hidden state $S_t^{\it suffix}$.

In addition, the embedding of the predicted suffix is not further fed into the hidden state of the next stem, because we think suffix information can provide little information for predicting the next stem from a linguistic perspective.

\subsection{Training}
During the training stage, the objective function $L$ consists of two components:
\begin{equation}
\label{objective}
    L=(1-\lambda)*L_{\it stem} + \lambda*L_{\it suffix},
\end{equation}
where:
\begin{equation}
\label{stem-cost}	
	L_{\it stem}=\sum_{i=1}^{n}p(y_i^{\it stem}|y_{i-1}^{\it stem},...,y_1^{\it stem},\textbf{x})
\end{equation}
and
\begin{equation}
\label{suffix-cost}
	L_{\it suffix} = \sum_{i=1}^{n}p(y_i^{\it suffix}|y_{i}^{\it stem},...,y_1^{\it stem},\textbf{x})
\end{equation}
$\lambda$ verifies from 0 to 1, and $\lambda$ can also be modeled in the whole architecture, which will be studied in our future work. In our experiments, we set $\lambda$ to 0.1 empirically. We use Adam \cite{Kingma2014Adam} as our optimizing function.

\subsection{Decoding}
Beam search is adopted as our decoding algorithm. At each time step, the search space can be infeasible large if we take all the combinations of stems and suffixes into consideration. So we use cube pruning \cite{huang2007forest} to obtain n-best candidates. First, the top $n$ stems with the highest scores are pushed to the stack. Then for each stem, we predict the top $n$ suffixes, which will result in $n$ complete candidates. The candidates will be inserted to a priority queue, which keeps records of the top $n$ complete candidates. After all the stems are expanded, the final n-best candidates are obtained.

\begin{table}[t]
    \begin{center}
        \begin{tabular}{p{7.5cm}}
        \hline
        \small{\textbf{Title}} \\
        \hline
        Amazing hot selling air scent machine \\
        Large capacity men backpack bags. \\
        Strap slash neck women pencil dress \\
        \hline
        \small{\textbf{Description}} \\
        \hline
        Along with tie shoulder straps, three-quarter sleeves. \\
        Compare the detail sizes with yours. \\
       	\hline
        \small{\textbf{Comment}} \\
        \hline
        I did not expect that the backpack is so happy. \\
        Thanks for the very quick shipping. \\
        I liked the dress. the quality is good. \\
        \hline
        \end{tabular}
    \end{center}
    \caption{Example of the e-commerce test set.}
    \label{tab:ecommerce-testset}
\end{table}

\begin{table*}[t]
    \begin{center}
        \begin{tabular}{l|cc|cc|ccc}
        \hline
        & \multicolumn{2}{c}{Vocabulary} & \multicolumn{2}{c}{Coverage} & \multicolumn{3}{c}{Test set}\\
        \hline
        Systems & Source & Target & Source & Target & News2014 & News2015 & News2016 \\
        \hline
        \hline
        RNN-based + Subword & 30K & 30K & 99.7\% & 97.0\% & 19.72(22.59) & 16.11 & 15.41\\
        Fully Character-based & 861 & 853 & 100\% & 100\% & 20.32(25.74) & 17.60 & 15.65\\
        RNN-based + \textbf{Suffix Prediction} & 30K & 30K & 99.7\% & 100\% & \textbf{21.30(26.22)} & \textbf{18.09} & \textbf{17.09}\\
        \hline
        Transformer + Subword & 30K & 30K & 99.7\% & 97.0\% & 23.18(26.39) & 18.66 & 18.31\\
        Transformer + \textbf{Suffix Prediction} & 30K & 30K & 99.7\% & 100\% & \textbf{24.41(29.14)} & \textbf{20.54} & \textbf{19.62}\\
        \hline
        \end{tabular}
    \end{center}
    \caption{Evaluation on the \textbf{news} domain: ``Subword'' refers to \citeauthor{sennrich2015neural} (\citeyear{sennrich2015neural}), ``Fully Character-based'' refers to \citeauthor{lee2016fully} (\citeyear{lee2016fully}), ``\textbf{Suffix Prediction}'' refers to our work. Scores in brackets are BLEU of \textbf{stem}, which means that the output sentence and reference are both transformed into stem sequence.}
    \label{tab:en2ru-bleu-news}
\end{table*}

\begin{table*}[t]
    \begin{center}
        \begin{tabular}{l|cc|cc|ccc}
        \hline
        & \multicolumn{2}{c}{Vocabulary} & \multicolumn{2}{c}{Coverage} & \multicolumn{3}{c}{Test set}\\
        \hline
        Systems & Source & Target & Source & Target & Title & Offer & Comments\\
        \hline
        \hline
        RNN-based + Subword & 45K & 45K & 99.8\% & 100\% & 17.52 & 29.78 & 33.29\\
        RNN-based + \textbf{Suffix Prediction} & 45K & 45K & 99.8\% & 100\% & \textbf{17.85} & \textbf{30.60} & \textbf{34.18}\\
        \hline
        \end{tabular}
    \end{center}
    \caption{Evaluation on the \textbf{e-commerce} domain: ``Subword'' refers to \citeauthor{sennrich2015neural} (\citeyear{sennrich2015neural}), ``\textbf{Suffix Prediction}'' refers to our work.}
    \label{tab:en2ru-bleu-ecomm}
\end{table*}

\begin{figure}[t]
\center
\includegraphics[width=0.45\textwidth]{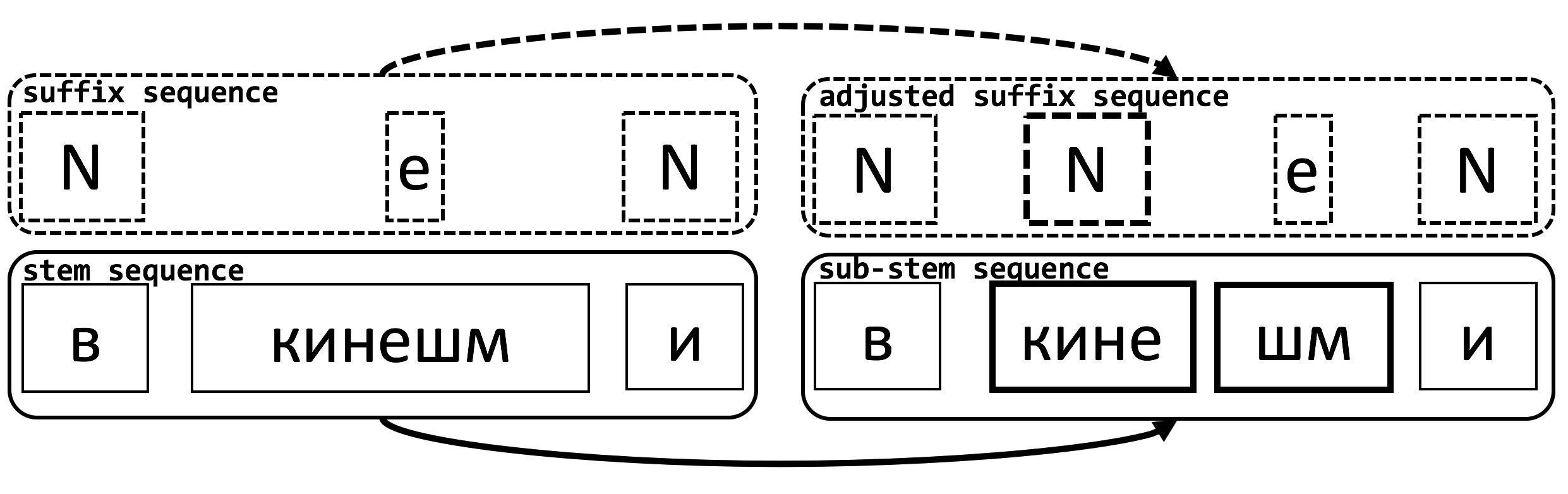}
\caption{Adjust the suffix sequence according to the ``sub-stem'' sequence.}
\label{fig:adjustinf}
\end{figure}

\section{Experiments}
We run our experiments on English to Russian (En-RU) data under two significantly different domain, namely the news domain and the e-commerce domain. We verify our method on both \textbf{RNN} based NMT architecture and \textbf{Transformer} based NMT architecture.

\subsection{Data}
\textbf{News} We select 5.3M sentences from the bilingual training corpus released by WMT2017 shared task on the news translation domain\footnote{http://www.statmt.org/wmt17/translation-task.html} as our training data. We use 3 test set, which are published by WMT2017 news translation task, namely ``News2014'', ``News2015'', ``News2016''.

\textbf{E-commerce} We collect 50M bilingual sentences as our training corpus:
\begin{itemize}
\item 10M sentences are crawled and automatic aligned from some international brand's English and Russian websites.
\item 20M are back translated corpus: First we crawled the Russian sentences from websites of certain Russian's Brands. Then translated them to English through a machine translation system trained on limited RU-EN corpus \cite{Sennrich2015Improving}.
\item The last 20M bilingual sentences are crawled from the web, and are not domain specific. 
\end{itemize}

We typically use the following 3 types of data as test set, which are named \textbf{title}, \textbf{description} and \textbf{comment}, these sentences are all extracted from e-commerce websites. \textbf{Title} are the goods' titles showed on a listing page when some buyers type in some keywords in a searching bar under an e-commerce website. \textbf{Description} refers to the information in a commodities' detail page. \textbf{Comment} include the review or feedback from some buyers. Example sentences are shown in Table \ref{tab:ecommerce-testset}. For each kind of test set, we randomly select 1K English sentences and translate it by human.

\textbf{Pre-Processing} Both the training set and the test set are lowercased, and some entity words appeared in the data are generalized into specific symbols, such as ``\_date\_'', ``\_time\_'', ``\_number\_''. When selecting our training data, we keep the sentences which has length between 1 to 30. We use a bilingual sentence scorer to discard some low-quality bilingual sentences. The scorer is simply trained under algorithm of IBM Model 1 \cite{brown1993mathematics} on a very large bilingual corpus.

\textbf{Target Side Word Stemming} We use snowball\footnote{http://snowball.tartarus.org/} to create stems from words. Because stem created from snowball is always a substring of the original word, we can obtain suffixes by simply applying a string cut operation. By applying snowball to a target side word sequence, we split a target side sentence into a \textbf{stem} sequence and a \textbf{suffix} sequence. The stemming accuracy of snowball is 83.3\% on our human labeled test set.

\textbf{Applying BPE to Target Side Stem Sequence} We also use the Byte-pair encoding (BPE algorithm) on the target side stem sequence, which will further reduce data sparsity. Some stems will be split into ``sub-stem'' units. The stem sequence is transferred to ``sub-stem'' sequence at this step. Suffix sequence should also be adjusted according to the ``sub-stem'' sequence simultaneously. More specifically, as shown in Figure \ref{fig:adjustinf}, if a stem is split into $n$ ``sub-stem'' units, then $n-1$ ``N'' (refers to ``N'' in Figure \ref{fig:stemsequence}) will be inserted into the suffix sequence, and these tags will be located in front of the suffix which is corresponding to the original complete stem. The sub-stem sequence and the adjusted suffix sequence are the final training corpus on target side.

\begin{figure*}[t]
\center
\includegraphics[width=1.0\textwidth]{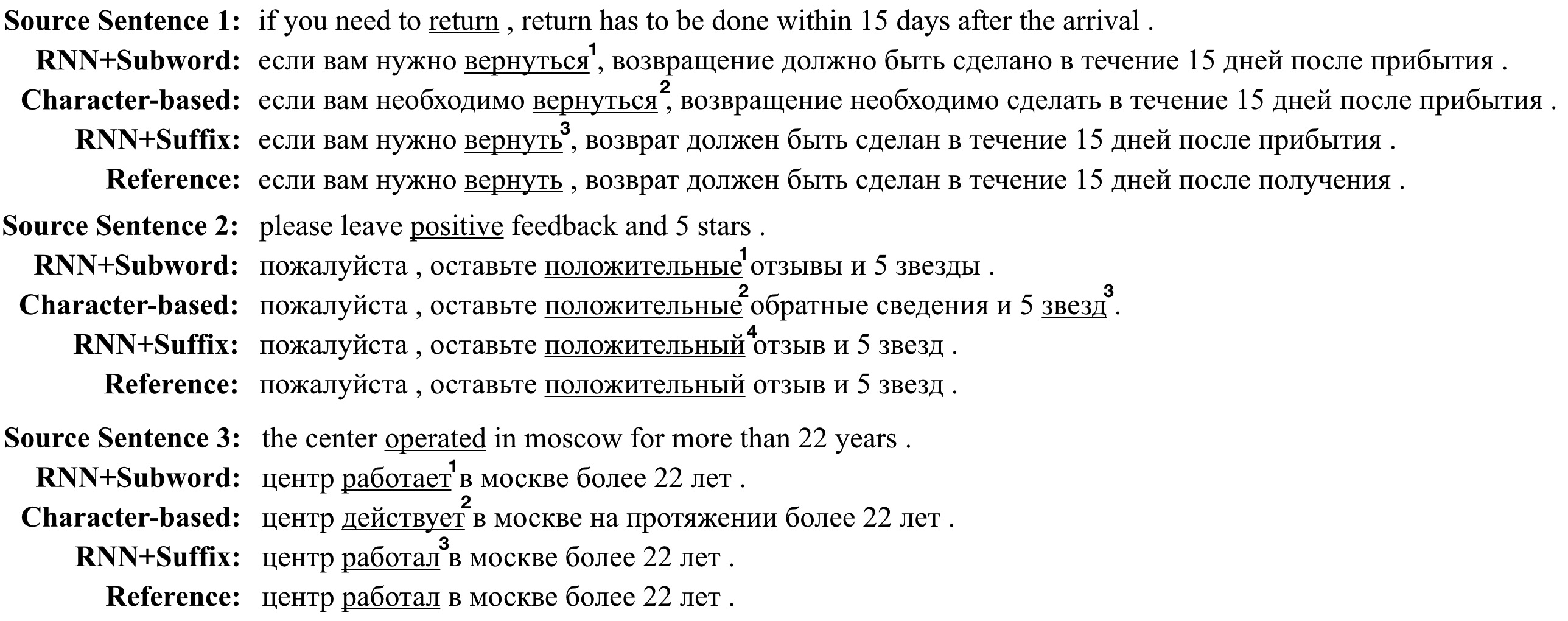}
\caption{``\textbf{RNN+Subword}'' refers to \citeauthor{sennrich2015neural} (\citeyear{sennrich2015neural}), ``\textbf{Character-based}'' refers to \citeauthor{lee2016fully} (\citeyear{lee2016fully}), ``\textbf{RNN+Suffix}'' refers to our work.}
\label{fig:sample}
\end{figure*}

\subsection{Baselines}
Our \textbf{RNN} and \textbf{Transformer} baseline systems utilize BPE \cite{sennrich2015neural} to transfer the original word sequence to subword sequence on both the source and the target sides, since the subword method had a stable improvement compared with word based system, especially on morphologically rich languages. 

Besides, we compared our system with a \textbf{fully character-based} baseline system, which is an implementation of \citeauthor{lee2016fully} (\citeyear{lee2016fully})'s work, and is available on github\footnote{https://github.com/nyu-dl/dl4mt-c2c}.

We limit the source and target vocabularies to the most frequent 30K tokens for both English and Russian. For news domain, about 99.7\% tokens are covered by the source side vocabulary, about 97.0\% target tokens are covered by the target side vocabulary.

\subsection{Our System}
For our system, the source token coverage is the same as the baselines. On the other hand, 100\% target tokens are covered by the target-side vocabulary, which consists of ``sub-stem'' units generated from target side stem sequence by applying BPE algorithm. There are totally 752 types of suffixes, which are calculated from the suffix sequences generated from target side sentences.

\subsection{Distributed Training}
For the experiments on the e-commerce domain, the training data is large. We use a distributed training framework for both the baseline system and our system. Training data are split into several parts, each being trained on a single worker node. A parameter server averages the model parameters from each worker node after every 100 training batchs and then synchronizes the averaged model to every worker node. Each worker continues with the training process based on the averaged model.

\subsection{Results and Analysis}
We use BLEU \cite{papineni:2002} as our evaluation metric. The performance of different systems are shown in Table \ref{tab:en2ru-bleu-news} and \ref{tab:en2ru-bleu-ecomm}. On both the news and e-commerce domains, our system performs better than baseline systems.

On news domain, the average improvement of our method is 1.75 and 0.97 BLEU score when implemented on RNN-based NMT, compared with subword \cite{sennrich2015neural} method and fully character-based \cite{lee2016fully} method, respectively. When implemented on Transformer \cite{DBLP:journals/corr/VaswaniSPUJGKP17}, average improvement is 1.47 BLEU compared with subword method. On the e-commerce domain, which use 50M sentences as training corpus, the average improvement of our method is 0.68 BLEU compared with the subword method.

We evaluate stem accuracies and suffix accuracies separately. For stem, we use BLEU as evaluation metric, Table \ref{tab:en2ru-bleu-news} shows stem BLEU of different methods on ``News2014'' test set, our method can gain significant improvement compared with baselines, since our method can reduce data sparsity better than baselines. Our method can effectively reduce suffix error, Figure \ref{fig:sample} gives some examples both on e-commerce and news domains:
\begin{itemize}
\item For the first sample, the suffix of the translation words (tagged by 1 and 2) from two different baseline systems means a reflexive verb, whose direct object is the same as its subject. In other words, a reflexive verb has the same semantic agent and patient. It is an incorrect translation according to the source meaning, because we can infer from the source sentence that the agent is a person and the patient is an object (some goods bought by a customer). In our system, the suffix of the translation word (tagged by 3) is correct. It represents an infinitive verb which may take objects, other complements and modifiers to form a verb phrase.
\item In the second sample, the translation word (tagged by 1) is not accurate, its suffix represents a plural form, but the correct form is singular, because the corresponding source word ``positive'' is singular form. Character-based system can correctly translate source word ``stars'' into a Russian word with plural form. However, the translation of ``positive'' (tagged by 2) is still with wrong form. Both the translation of ``positive'' and ``stars'' from our system are with the correct forms. 
\item In the third sample, the translation word tagged by 3 represents past tense; However, the translation words tagged by 1 and 2 represent present tense. Our system successfully predicted the tense moods.
\end{itemize}

\section{Conclusion}
We proposed a simple but effective method to improve English-Russian NMT, for which a morphologically rich language is on the target side. We take a two-step approach in the decoder. At each step, a stem is first generated, then its suffix is generated. We empirically compared our method with two previous methods (namely subword and fully character-based), which can also to some extent address our problem. Our method gives an improvement on two encoder-decoder NMT architectures on two domains. To our knowledge, we are the first to explicitly model suffix for morphologically-rich target translation.

\section{Acknowledgments}
We thank the anonymous reviewers for their detailed and constructed comments. Yue Zhang and Min Zhang are the corresponding authors. The research work is supported by the National Natural Science Foundation of China (61525205, 61432013, 61373095). Thanks for Xiaoqing Li, Heng Yu and Zhdanova Liubov for their useful discussion.
\bibliographystyle{aaai}
\bibliography{draft}

\begin{thebibliography}{}

\bibitem[\protect\citeauthoryear{Ba, Kiros, and Hinton}{2016}]{ba2016layer}
Ba, J.~L.; Kiros, J.~R.; and Hinton, G.~E.
\newblock 2016.
\newblock Layer normalization.
\newblock {\em arXiv preprint arXiv:1607.06450}.

\bibitem[\protect\citeauthoryear{Bahdanau, Cho, and
  Bengio}{2014}]{bahdanau2014neural}
Bahdanau, D.; Cho, K.; and Bengio, Y.
\newblock 2014.
\newblock Neural machine translation by jointly learning to align and
  translate.
\newblock {\em arXiv preprint arXiv:1409.0473}.

\bibitem[\protect\citeauthoryear{Brown \bgroup et al\mbox.\egroup
  }{1993}]{brown1993mathematics}
Brown, P.~F.; Pietra, V. J.~D.; Pietra, S. A.~D.; and Mercer, R.~L.
\newblock 1993.
\newblock The mathematics of statistical machine translation: Parameter
  estimation.
\newblock {\em Computational Linguistics} 19(2):263--311.

\bibitem[\protect\citeauthoryear{Chahuneau \bgroup et al\mbox.\egroup
  }{2013}]{chahuneau2013translating}
Chahuneau, V.; Schlinger, E.; Smith, N.~A.; and Dyer, C.
\newblock 2013.
\newblock Translating into morphologically rich languages with synthetic
  phrases.
\newblock In {\em Prague Bulletin of Mathematical Linguistics}, volume 100,
  51--62.

\bibitem[\protect\citeauthoryear{Cho \bgroup et al\mbox.\egroup
  }{2014a}]{cho2014properties}
Cho, K.; Van~Merri{\"e}nboer, B.; Bahdanau, D.; and Bengio, Y.
\newblock 2014a.
\newblock On the properties of neural machine translation: Encoder-decoder
  approaches.
\newblock {\em arXiv preprint arXiv:1409.1259}.

\bibitem[\protect\citeauthoryear{Cho \bgroup et al\mbox.\egroup
  }{2014b}]{cho2014learning}
Cho, K.; Van~Merri{\"e}nboer, B.; Gulcehre, C.; Bahdanau, D.; Bougares, F.;
  Schwenk, H.; and Bengio, Y.
\newblock 2014b.
\newblock Learning phrase representations using rnn encoder-decoder for
  statistical machine translation.
\newblock {\em arXiv preprint arXiv:1406.1078}.

\bibitem[\protect\citeauthoryear{Goodfellow \bgroup et al\mbox.\egroup
  }{2013}]{goodfellow2013maxout}
Goodfellow, I.~J.; Warde-Farley, D.; Mirza, M.; Courville, A.; and Bengio, Y.
\newblock 2013.
\newblock Maxout networks.
\newblock {\em arXiv preprint arXiv:1302.4389}.

\bibitem[\protect\citeauthoryear{Hochreiter and
  Schmidhuber}{1997}]{hochreiter1997long}
Hochreiter, S., and Schmidhuber, J.
\newblock 1997.
\newblock Long short-term memory.
\newblock {\em Neural Computation} 9(8):1735--1780.

\bibitem[\protect\citeauthoryear{Huang and Chiang}{2007}]{huang2007forest}
Huang, L., and Chiang, D.
\newblock 2007.
\newblock Forest rescoring: Faster decoding with integrated language models.
\newblock In {\em ACL}, volume~45,  144.

\bibitem[\protect\citeauthoryear{Jean \bgroup et al\mbox.\egroup
  }{2014}]{jean2014using}
Jean, S.; Cho, K.; Memisevic, R.; and Bengio, Y.
\newblock 2014.
\newblock On using very large target vocabulary for neural machine translation.
\newblock {\em arXiv preprint arXiv:1412.2007}.

\bibitem[\protect\citeauthoryear{Kingma and Ba}{2014}]{Kingma2014Adam}
Kingma, D.~P., and Ba, J.
\newblock 2014.
\newblock Adam: A method for stochastic optimization.
\newblock {\em Computer Science}.

\bibitem[\protect\citeauthoryear{Koehn and Hoang}{2007}]{koehn2007factored}
Koehn, P., and Hoang, H.
\newblock 2007.
\newblock Factored translation models.
\newblock In {\em EMNLP-CoNLL},  868--876.

\bibitem[\protect\citeauthoryear{Lee, Cho, and Hofmann}{2016}]{lee2016fully}
Lee, J.; Cho, K.; and Hofmann, T.
\newblock 2016.
\newblock Fully character-level neural machine translation without explicit
  segmentation.
\newblock {\em arXiv preprint arXiv:1610.03017}.

\bibitem[\protect\citeauthoryear{Luong and Manning}{2016}]{luong2016achieving}
Luong, M.-T., and Manning, C.~D.
\newblock 2016.
\newblock Achieving open vocabulary neural machine translation with hybrid
  word-character models.
\newblock {\em arXiv preprint arXiv:1604.00788}.

\bibitem[\protect\citeauthoryear{Luong \bgroup et al\mbox.\egroup
  }{2014}]{Luong2014Addressing}
Luong, M.~T.; Sutskever, I.; Le, Q.~V.; Vinyals, O.; and Zaremba, W.
\newblock 2014.
\newblock Addressing the rare word problem in neural machine translation.
\newblock {\em Bulletin of University of Agricultural Sciences and Veterinary
  Medicine Cluj-Napoca. Veterinary Medicine} 27(2):82--86.

\bibitem[\protect\citeauthoryear{Mi, Wang, and
  Ittycheriah}{2016}]{mi2016vocabulary}
Mi, H.; Wang, Z.; and Ittycheriah, A.
\newblock 2016.
\newblock Vocabulary manipulation for neural machine translation.
\newblock {\em arXiv preprint arXiv:1605.03209}.

\bibitem[\protect\citeauthoryear{Papineni \bgroup et al\mbox.\egroup
  }{2002}]{papineni:2002}
Papineni, K.; Roukos, S.; Ward, T.; and Zhu, W.-J.
\newblock 2002.
\newblock Bleu: a method for automatic evaluation of machine translation.
\newblock In {\em ACL},  311--318.

\bibitem[\protect\citeauthoryear{Sennrich and
  Haddow}{2016}]{sennrich2016linguistic}
Sennrich, R., and Haddow, B.
\newblock 2016.
\newblock Linguistic input features improve neural machine translation.
\newblock {\em arXiv preprint arXiv:1606.02892}.

\bibitem[\protect\citeauthoryear{Sennrich, Haddow, and
  Birch}{2015a}]{Sennrich2015Improving}
Sennrich, R.; Haddow, B.; and Birch, A.
\newblock 2015a.
\newblock Improving neural machine translation models with monolingual data.
\newblock {\em Computer Science}.

\bibitem[\protect\citeauthoryear{Sennrich, Haddow, and
  Birch}{2015b}]{sennrich2015neural}
Sennrich, R.; Haddow, B.; and Birch, A.
\newblock 2015b.
\newblock Neural machine translation of rare words with subword units.
\newblock {\em arXiv preprint arXiv:1508.07909}.

\bibitem[\protect\citeauthoryear{Song \bgroup et al\mbox.\egroup
  }{2014}]{song2014joint}
Song, L.; Zhang, Y.; Song, K.; and Liu, Q.
\newblock 2014.
\newblock Joint morphological generation and syntactic linearization.
\newblock In {\em AAAI},  1522--1528.

\bibitem[\protect\citeauthoryear{Srivastava \bgroup et al\mbox.\egroup
  }{2014}]{srivastava2014dropout}
Srivastava, N.; Hinton, G.~E.; Krizhevsky, A.; Sutskever, I.; and
  Salakhutdinov, R.
\newblock 2014.
\newblock Dropout: a simple way to prevent neural networks from overfitting.
\newblock {\em Journal of machine learning research} 15(1):1929--1958.

\bibitem[\protect\citeauthoryear{Tamchyna, Marco, and
  Fraser}{2017}]{Ale2017Modeling}
Tamchyna, A.; Marco, M.~W.; and Fraser, A.
\newblock 2017.
\newblock Modeling target-side inflection in neural machine translation.
\newblock {\em WMT}.

\bibitem[\protect\citeauthoryear{Toutanova, Suzuki, and
  Ruopp}{2010}]{Toutanova2010Applying}
Toutanova, K.; Suzuki, H.; and Ruopp, A.
\newblock 2010.
\newblock Applying morphology generation models to machine translation.
\newblock In {\em ACL},  514--522.

\bibitem[\protect\citeauthoryear{Tran, Bisazza, and
  Monz}{2015}]{tran2015distributed}
Tran, K.; Bisazza, A.; and Monz, C.
\newblock 2015.
\newblock A distributed inflection model for translating into morphologically
  rich languages.
\newblock {\em Proceedings of MT Summit XV}  145.

\bibitem[\protect\citeauthoryear{Vaswani \bgroup et al\mbox.\egroup
  }{2017}]{DBLP:journals/corr/VaswaniSPUJGKP17}
Vaswani, A.; Shazeer, N.; Parmar, N.; Uszkoreit, J.; Jones, L.; Gomez, A.~N.;
  Kaiser, L.; and Polosukhin, I.
\newblock 2017.
\newblock Attention is all you need.
\newblock {\em CoRR} abs/1706.03762.

\bibitem[\protect\citeauthoryear{Zens, Och, and Ney}{2002}]{Zens2002Phrase}
Zens, R.; Och, F.~J.; and Ney, H.
\newblock 2002.
\newblock Phrase-based statistical machine translation.
\newblock {\em Lecture Notes in Computer Science} 11(2):18--32.

\end{thebibliography}

\end{document}